\documentclass[letterpaper, 10 pt, conference]{ieeeconf}  

\IEEEoverridecommandlockouts                              

\usepackage{graphics} 
\usepackage{epsfig} 
\usepackage{mathptmx} 
\usepackage{times} 
\usepackage{amsmath} 
\usepackage{amssymb}  
\usepackage{graphicx}
\usepackage{tabularx,booktabs}
\usepackage{hyperref}
\usepackage{cite}
\usepackage{comment}
\usepackage{booktabs}
\usepackage{alphabeta} 
\usepackage{authblk}
\usepackage[font=small,skip=0pt]{caption}

\title{\LARGE \bf
Single-Fiber Optical Frequency Domain Reflectometry  Shape Sensing of Continuum Manipulators with Planar Bending
}
\author{Mobina Tavangarifard, Wendy Rodriguez Ovalle, and Farshid Alambeigi \IEEEmembership{Member, IEEE}
\thanks{This work is supported by the National Institute Of Biomedical Imaging and Bioengineering of the National Institutes of Health under Award Number R21EB030796.}
\thanks{$^{1}$M.~Tavangarifard, W. R. Ovalle, and F.~Alambeigi are with the Walker Department of Mechanical Engineering and the Texas Robotics  at the University of Texas at Austin, Austin, TX, 78712, USA. Email: mt39884@my.utexas.edu; wendyrovalle@utexas.edu,   farshid.alambeigi@austin.utexas.edu}.}

\begin{document}

\maketitle
\thispagestyle{empty}
\pagestyle{empty}

\begin{abstract}
To address the  challenges associated with shape sensing  of continuum manipulators (CMs) using Fiber Bragg Grating (FBG) optical fibers, we  feature a unique shape sensing assembly  utilizing solely a single \textit{Optical Frequency Domain Reflectometry (OFDR)} fiber attached to a flat nitinol wire (NiTi). Integrating this easy-to-manufacture unique sensor with a long and soft  CM with 170 mm length, we performed different experiments to evaluate its shape reconstruction ability.   Results demonstrate phenomenal shape reconstruction accuracy for both C-shape ($<$ 2 mm tip error, $<$ 1.2 mm shape error) and J-shape ($<$ 3.4 mm tip error, $<$ 2.3 mm shape error) experiments. 


\end{abstract}

\section{INTRODUCTION}

Instead of using rigid instrumentation  in minimally invasive surgeries  that limit access and dexterity of clinicians, continuum manipulators (CMs) have been proposed to provide more flexibility and access within confined  and tortuous paths of anatomical spaces \cite{shi2016shape}. 
While the use of CMs have been deemed favorable for the traits mentioned before, there are ongoing challenges with using CMs, such as complex modeling, sensing, and control of these robots \cite{liuImpact2022,liuTmech2022}. 
 Particularly, having an appropriate integrated sensing modality and pertinent algorithm to estimate their shape and end effector position at any time would be of Paramount importance for accurate control of these robots \cite{alambeigi2019scade,shi2016shape}.  

While kinematic/dynamic models have been used to estimate the shape of CMs, model-based approaches often result in significant errors due to highly nonlinear and entangled deformation behavior of CMs and various complex phenomena such as  hysteresis and tension loss \cite{shi2016shape,da2020challenges,liuTmech2022}. Alternative methods to conduct shape sensing (SS) in real-time include but not limited to the use of electromagnetic tracking,  intraoperative fluoroscopic imaging, and, Fiber Bragg Grating (FBG) optical sensing  \cite{shi2016shape}. Among these  electromagnetic (EM) tracking typically suffers from low SS accuracy  specially in in a relatively large workspace and in the presence of ferromagnetic instruments. The fluoroscopic imaging also is not very favorable and cannot be used continuously due to the radiation and safety issues created for the clinicians and patients during the procedure \cite{shi2016shape,Sorriento2020EM}.
\begin{figure}[t]
    \centering
    \includegraphics[width=\columnwidth]{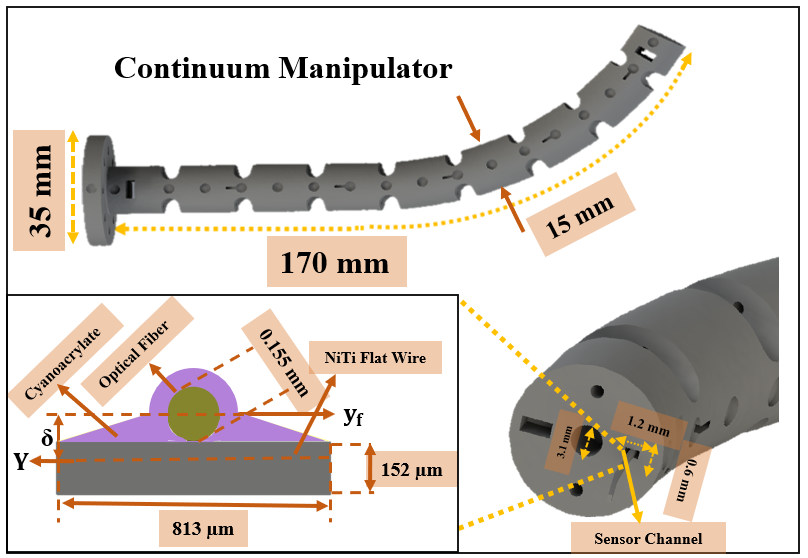}
    \caption{An overview of the used the sensorized  CM with the OFDR-based SSA. Figure also shows the SSA structure modeled as a composite beam. } 
    \label{fig:OFDR-SSA}
\end{figure}
On the other hand, FBGs offer various benefits such as minimal footprint and seamless integration with rigid and flexible instruments, relatively large sensing length, biocompatibility, and dynamic SS \cite{alambeigi2019scade}. Despite these advantages, FBG-based SS also suffers from  the time-consuming, labor-intensive, and expensive fabrication procedure (e.g., infrared femtosecond
laser cutting \cite{sefati2019fbg,monet2020high})  resulting in a low SS accuracy \cite{sefati2019fbg,monet2020high}. Moreover,  the limited number of Gratings nodes or measurement points  along  an FBG fiber directly affects the spatial resolution and results in a poor SS accuracy. Moreover, increasing the number of FBG nodes cannot completely address this poor accuracy and may increase the wavelength shifts interfering with each other,
especially during high curvature bending \cite{hanley2023trade,monet2020high}. 

To address the poor accuracy of FBG-based SS, several sensor fusion algorithms have been suggested in the literature. For example, Alambeigi et al. \cite{alambeigi2019scade} fused the low accuracy and high frequency estimations of FBG SS with high accuracy and low frequency  fluoroscopic estimations to improve the SS accuracy of a tendon-driven CM.   Moreover, Densai et al. \cite{denasi2018observer} introduced   an observer and Kalman filter to fuse FBG SS and ultrasound imaging to better track the tip of a magnetically actuated catheter. Also,  in \cite{shi2014simultaneous}, FBG SS was used with   electromagnetic tracking and ultrasound imaging for shape sensing of a catheter.  

As an alternative SS approach for CMs, Optical Frequency Domain Reflectometry-based (OFDR-based) SS has also recently been introduced  \cite{parent2018intra,parent2017enhancement,monet2020high}. Despite FBG-based SS that uses   \textit{expensive} FBG  fibers that suffer from complex fabrication procedure, in OFDR-based SS, \textit{regular} optical fibers can be used in a simpler shape sensing assembly \cite{nguyen2022toward}. Additionally, a benefit of using OFDR-based sensors over FBG-based sensors is the presence of \textit{continuous} strain measurements and \textit{distributed} SS along the entire length of the fiber to provide more information on the status of the sensor \cite{froggatt1998high}. In a recent study \cite{monet2020high}, authors compared the performance of the OFDR- and  FBG-based SS in a tendon-driven CM. Nevertheless, similar to the FBG-based SS, the proposed manufacturing and assembly process   is very costly and difficult. Also, multiple fibers have been used in the proposed sensor.  manufacturing and assembly procedures as well as utilizing multiple fibers in the SSA   are still necessary problems to be addressed. 

To address the aforementioned fabrication challenges of FBG-based and existing  OFDR-based SS, in \cite{nguyen2022toward}, we proposed a sensor design and Shape Sensing Assembly (SSA) using a flat wire and a single OFDR fiber.  We also demonstrated the calibration procedure for this sensor. However, in that study, we did not evaluate the SS performance of this SSA inside a CM and in different bending situations. Therefore, in this paper and as our main \textit{contributions}, we leveraged the design of this OFDR-based  SSA, and evaluated its SS performance after integrating that inside an additively manufactured soft CM (Fig. \ref{fig:OFDR-SSA}). To thoroughly assess the performance of the SSA, we experimentally evaluated its SS ability when the CM experienced  C- and J-shapes with different curvatures. 

\section{Methodology}

\subsection{Sensor Design and Assembly} 

In \cite{nguyen2022toward}, we proposed an OFDR-SSA by attaching a single distributed optical fiber to a flat wire with a rectangular cross-section, as illustrated in Fig. \ref{fig:OFDR-SSA}. Importantly, we demonstrated that employing a flat NiTi wire -- as opposed to a common round one used in the literature (e.g.,  \cite{liu2015large, sefati2019fbg}) -- can dramatically  simplify the SSA manufacturing process, making it quicker, more affordable, and more consistent. Of note, typically, a triangular pattern  is used as the SSA configuration in which either one optical fiber is connected to two NiTi wires with a circular cross-section (e.g., \cite{liu2015large}), or a single NiTi wire with three laser-cut grooves on its circular cross-section, to which three optical fibers are attached  \cite{sefati2019fbg}. The following briefly describes the design and fabrication process of this OFDR-based SSA. More details about this SSA can be found in   \cite{nguyen2022toward}.
 \begin{figure}[t!]
    \centering
    \includegraphics[width=\columnwidth]{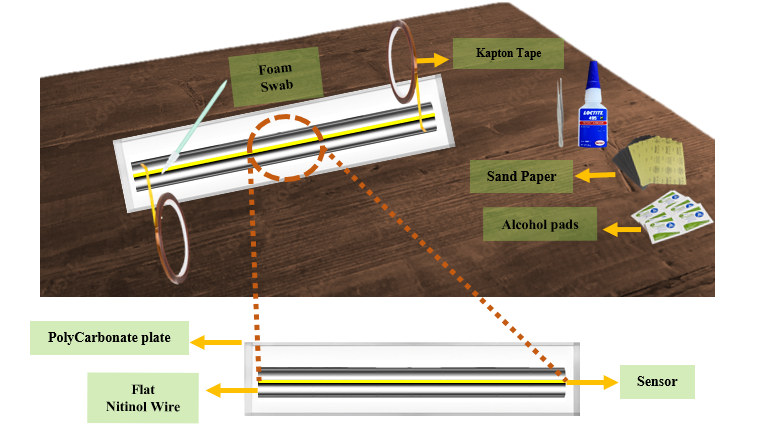}
    \caption{Fabrication setup used to mount the OFDR fiber on a flat NiTi wire.  }
    \label{fig: Fabrication Setup}
\end{figure}
\subsubsection{SSA Geometry Design}
  As the first step to designing the SSA, we first need to find the appropriate flat NiTi wire size to use for the OFDR fiber. Of note, the SSA geometry and particularly its neutral plane \( \overline{Y} \) directly impacts the maximum bending radius \( \rho_{\text{max}} \), the maximum allowable strain \( \epsilon_{\text{max}} \), and therefore the sensitivity of the sensor \cite{liu2015shape}.
To optimally design the SSA geometry,  we first modeled it as a composite beam shown in Fig. \ref{fig:OFDR-SSA}. The overall neutral plane of the SSA,  \( \overline{Y} \), can be calculated  knowing  the Young Modulus $E$ and the cross-section $A$ of the optical fiber (f) and NiTi wire (w). Moreover, sensor bias $\delta$ can be calculated knowing the radius of the used optical fiber (HD65, Luna Innovations Inc.) with $r_{f}=77.5$ $\mu m$ and its  maximum allowable strain $\epsilon_{max}=$1\% as follows:
\begin{equation}
\label{eq:1}
\overline{Y} = \frac{\sum_{j} \frac{E_{j}}{E_{f}} A_{j} \overline{y}_{j}}{\sum_{j} \frac{E_{j}}{E_{f}} \overline{y}_{j}}, \quad \text{where } j = (w, f)\quad 
\end{equation}
\begin{equation}
\label{eq:2}
\delta = \overline{y}_{f} - \overline{Y}
\end{equation}
where \( A_{j} \) represents the cross section area and \( \overline{y}_{j} \) is the  neutral plane of  fiber or wire. Also,  $E_{n}=75$ GPa and $E_{f}=4.81$ GPa are Young's modulus for the  NiTi wire, and  the optical fiber, respectively.   

Next, given known OFDR fiber (HD65, Luna Innovations Inc.) with maximum allowable bending radius $\rho_{max}=r_{f} / \epsilon_{max}$,  we   used (\ref{eq:1}) and (\ref{eq:2}) and  performed a search for a wide range of possible sizes for the NiTi wire  with a width ranging from 0.5 to 1 mm and heights from 0 to 0.5 mm to maximize  the sensor's sensitivity or bias. Considering the size of the considered CM with 15 mm diameter and a 1.2 mm $\times$ 0.6 mm rectangular SSA channel    (Fig. \ref{fig:OFDR-SSA}),  we chose a flat NiTi wire (Kellogg’s Research Labs) with dimensions of 0.152 mm $\times$ 0.813 mm. The selected  wire resulted in the sensor bias  $\delta=0.079$ mm \cite{nguyen2022toward}. Of note, these dimensions  can be readily scaled and optimized based on the application and size of the CM to maximize sensitivity of the sensor.

\subsubsection{Fabrication Process}
As conceptually shown in Fig. \ref{fig: Fabrication Setup}, the proposed manufacturing of SSA started with affixing a flat polycarbonate (PC) plate to a table, followed by taping the selected flat NiTi to the PC sheet using Kapton tape (Tapes Master Inc.). Before attachment, the NiTi wire was sanded using $400$ and $600$-grit sandpaper to achieve a smooth surface finish. The sanding process involved gently abrading the surface of the wire, and then, the surface of the NiTi wire was wiped clean using a lint-free cloth alcohol pad to remove any residual dust. Next, the fiber was placed in the middle of the NiTi wire along its length. Then, using a swab foam, a small amount of Cyanoacrylate adhesive (Loctite 495) was applied along the last $200$ mm of the end of the fiber. The adhesive was allowed to cure for $24$ hours. After curing, the Kapton tape was removed, and the SSA was ready to be implemented.

\subsection{SSA Calibration Procedure} \label{sec:cal}
\indent Before beginning the SS and reconstruction process, the relationship between the continuous strain along the SSA length and the curvature of the sensor needs to be found through a calibration process. To this end and as shown in Fig. \ref{fig:C and J Shape Jigs}a,  we first printed a calibration jig using stereo-lithography (SLA) 3D printing (Form3, Formlabs Inc) with ten constant curvatures slots    ranging from $0$ to $16.66$ m\textsuperscript{-1}. The process started with placing the fabricated SSA -- shown in  Fig. \ref{fig:C and J Shape Jigs}a -- inside the slots of the jig   during data collection. The strain data was collected by an OFDR-interrogator (ODiSI 6000 Series, Luna Innovations Inc) with a frequency of $62.5$ Hz and spatial resolution of $1.3$ mm. This procedure is repeated three times for each slot with a $5$-minute relaxation time between different curvatures and $2$ hours between each trial to prevent hysteresis effects.

\subsection {Continuum Manipulator Design and Fabrication}
  Fig. \ref{fig:OFDR-SSA} illustrates the designed and fabricated CM for integrating the fabricated  SSA within its internal sensor channel and performing SS procedures  in different curvatures. As shown, the   robot has lateral flexural patterns to constrain the manipulator to planar bending and increase its flexibility. It also has a 170 mm length with 15 mm outer and 3.1 mm inner diameters. To place the SSA, a rectangular channel ($0.6$ x $1.2$ mm) has been considered along the its length.  To fabricate the flexible CM, we used a PolyJet additive manufacturing process and Stratasys J750 printer with  TangoBlack material.  

\vspace*{-1mm}
\subsection{Shape Sensing Procedure and Experiments}
\begin{figure}[t]
    \centering
    \includegraphics[width=\columnwidth]{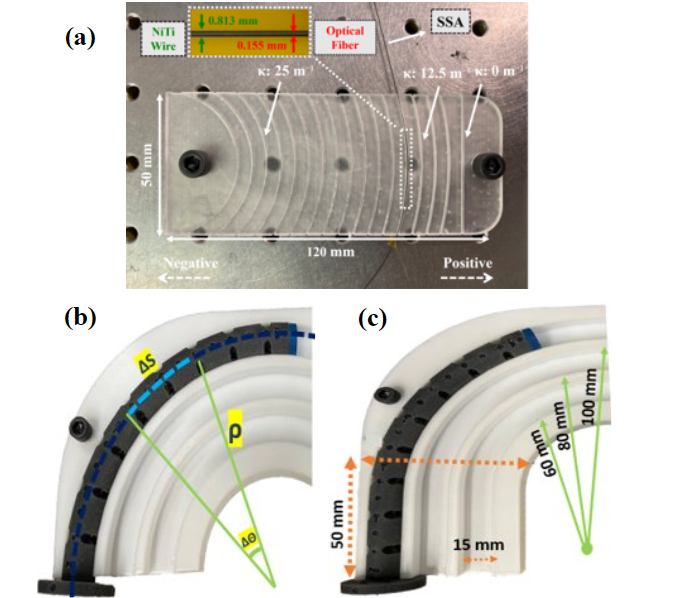}
    \caption{ (a) Calibration jig used to calibrate the SSA, (b) C-shape SS jig, and (c) J-shape SS jig  used to perform SS procedure.}
    \label{fig:C and J Shape Jigs}
\end{figure}

To perform the SS procedure using the sensorized CM (shown in Fig. \ref{fig:OFDR-SSA}), we first designed and 3D printed (Raise3D printer, PLA) two different SS jigs. The first jig contained three different semi-circle slots (Fig. \ref{fig:C and J Shape Jigs}b), while the second jig contained three different J-shape slots (Fig. \ref{fig:C and J Shape Jigs}c). Each  jig contained three different radius of curvature with $100$, $80$, and $60$ mm in radii, respectively, with a $15$ mm width and $9$ mm depth slot to hold the CM. The J-shaped jig also contained a straight $50$ mm long segment connected to each curvature. To perform experiments, as shown in the figure, the sensorized CM inserted into each slot of the jigs and the obtained strains where measured along the SSA. Three trials were conducted for each curvature of each jig with a $10$-minute resting period in between to ensure that the sensor returned to its original shape and minimized residual effects.
\begin{figure*}[t]
  \centering
     \includegraphics[width= \textwidth]{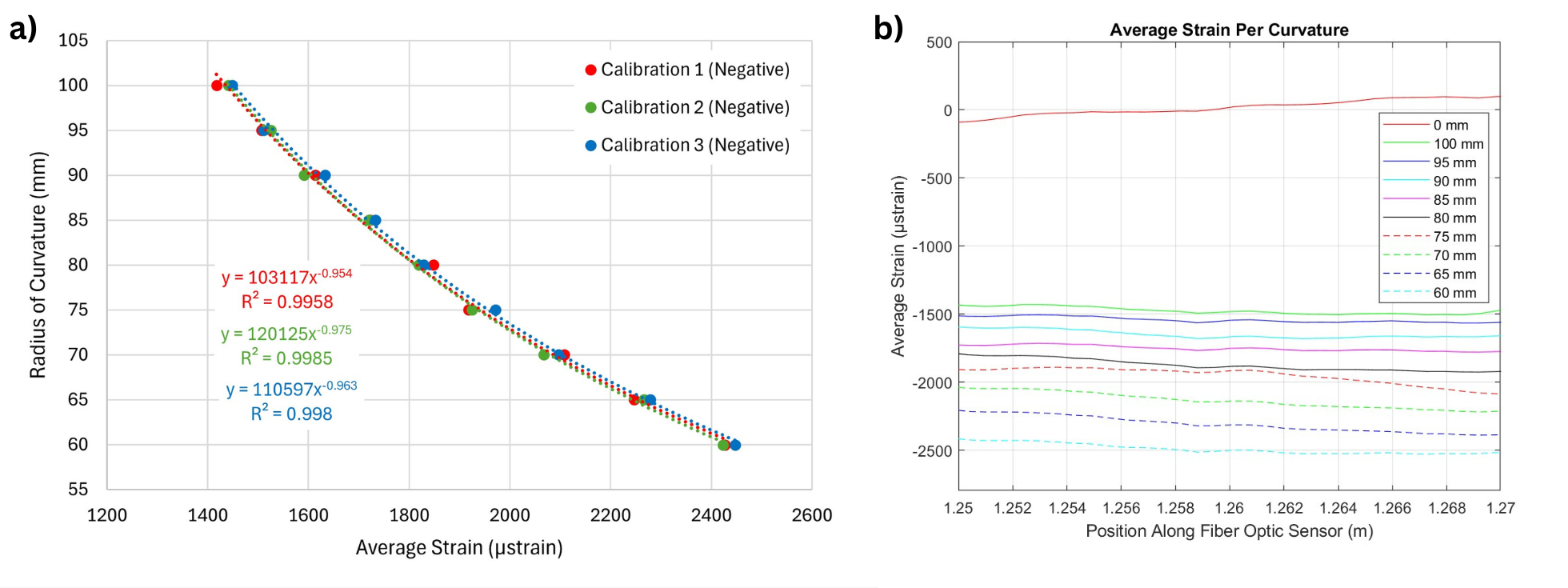}
      \caption{(a) Strain-curvature plot in the negative bending direction; (b) Average calibration strain measurements along the OFDR SSA for ten considered curves.}
      \label{fig:Calibration Results}
\end{figure*}

After obtaining the strain values along the SSA and and converting them to the curvatures ($\kappa$) using the performed calibration procedure and the strain-curvature relationship, we can use the following equations to reconstruct the shape of SSA by calculating the 2D position ($x$ and $y$) of each point along its length:   \cite{liu2015shape}:
\begin{equation}\label{theta}
\Delta\theta = \kappa\times\Delta s  ,  \kappa = \frac{1}{\rho} \tag{4}
\end{equation}
\begin{equation}\label{deltaX}
\Delta x = cos\theta (s) \times\Delta s, \tag{5}
\end{equation}
\begin{equation}\label{deltaY}
\Delta y = sin\theta (s)\times\Delta s \tag{6}
\end{equation}
where, as shown in Fig. \ref{fig:C and J Shape Jigs}, $\rho$ is the radius of the curvature in the used jigs, and $\Delta$s and $\Delta \theta$s represent the linear and angular distance between two points along the SSA's arc length $s$.   

Next, using (\ref{theta})-(\ref{deltaY}), the following process can be used to conduct shape reconstruction:

\indent \textbf{Step 1:} Using the strains measured by SSA, find the curvature along the length of the optical fiber using the strain-curvature relationship found through calibration procedure. 

\indent \textbf{Step 2:}  Calculate the distance ($\Delta$s)  between each point measured along the length of the  SSA using an arc length estimation described in (\ref{theta}).

\indent \textbf{Step 3:} Using the calculated curvature ($\kappa$) and distance between points ($\Delta$s), use (\ref{deltaX}) and (\ref{deltaY}) to solve for each successive point's X and Y coordinates along the SSA.

\indent \textbf{Step 4:} Plot and connect all calculated coordinates to form the reconstructed shape of the CM.
\subsection{Evaluation Metrics}
After performing shape reconstruction, the calculated coordinates of the SSA  are compared to the expected position for each point obtained from the known shape of each curvature. With this comparison, the CM's tip position error, shape reconstruction error, and average area error between the reconstructed shape and expected shape of the CM constrained in each jig  can be calculated. Of note, the tip position error measures the distance between the endpoint of the SSA and its expected position. The shape error measures the average error between the calculated and expected points for each point along the SSA.  Lastly, the average area between the estimated shape reconstruction and the expected shape is measured. Of note, an ideal shape reconstruction results in a zero area between the two curves. Of note, for ease of presentation, from this section forward, the curve with a radius of $100$ mm is referred to as C1, the curve with a radius of $80$ mm is referred to as C2, and the curve with a radius of $60$ mm is referred to as C3. The same simplification will be applied to the J-shaped curvatures based on the curved portion of their shape: the curvature with a $100$ mm radius will be referred to as J1, the curvature with a $80$ mm radius will be referred to as J2, and the curvature with a $60$ mm radius will be referred to as J3.

\section{Results }
\subsection{Calibration Results}
\indent Following the calibration procedure described in Section \ref{sec:cal}, the following radii were considered for the calibration procedure: $0$ mm, $100$ mm, $95$ mm, $90$ mm, $85$ mm, $80$ mm, $75$ mm, $70$ mm, $75$ mm, and $60$ mm, with $0$ mm representing a straight part h or zero curvature.   Figure \ref{fig:Calibration Results}a shows the obtained relationship between the strain and curvature for each trial. Also,  Fig. \ref{fig:Calibration Results}b shows the measured average strain of the three trials for all ten curvatures. Using all three calibration trials, the following relationship between strain and curvature was found using the average strains per curvature:
\begin{equation} \label{eq:rho}
\rho_{SSA} = 126099.3715 \times \epsilon_{SSA}^{-0.97984} \tag{7}
\end{equation}
where $\rho_{SSA}$  is the bending curvature of SSA in millimeters and $\epsilon_{SSA}$  is the measured strain of SSA in micro-strain units.

The relationship found in (\ref{eq:rho}) can now be used to calculate the curvature of other strains obtained during our C-shape and J-shape experiments. 

\begin{figure}[t]
    \centering
    \includegraphics[width=\columnwidth]{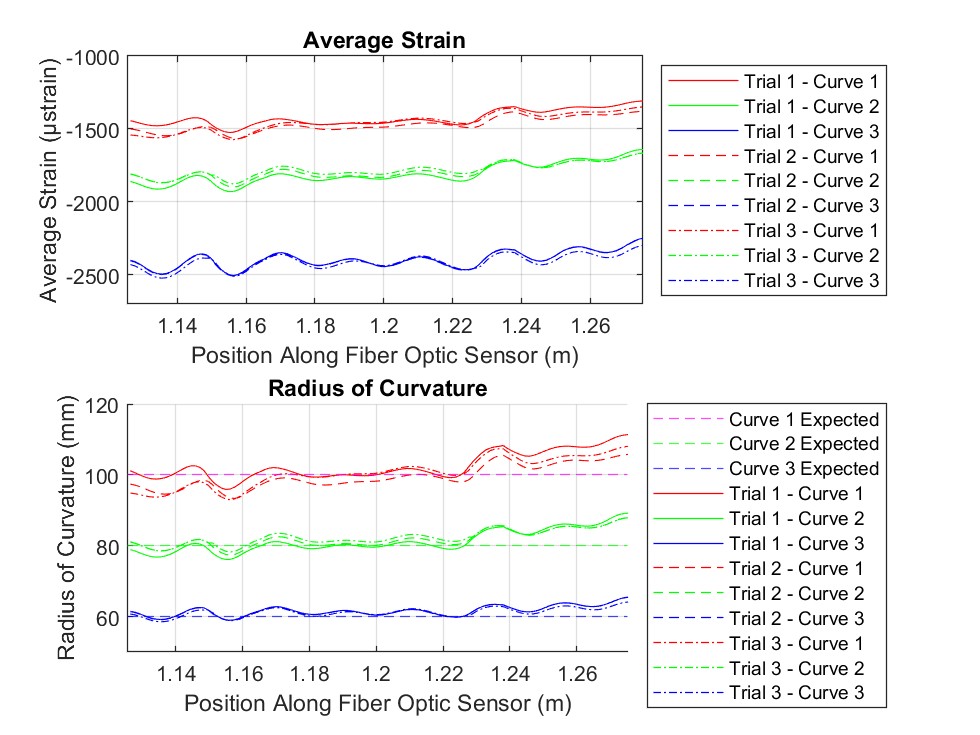}
    \caption{C-shape bending strain and radius of curvature along the length of the SSA. }
    \label{fig: C-Shape Strain and Curvature Plots}
\end{figure}
\begin{figure}[t]
    \centering
    \includegraphics[width=\columnwidth]{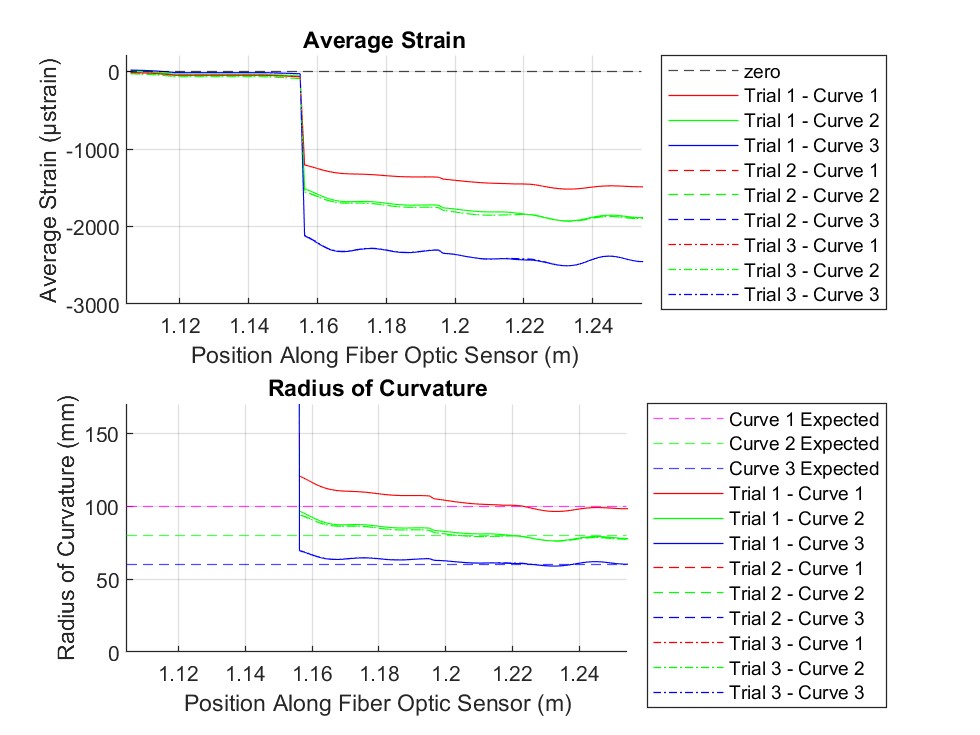}
    \caption{J-shape bending strain and radius of curvature along the length of the SSA. }
    \label{fig: J-Shape Strain and Curvature Plots}
\end{figure}


\begin{figure}[t]
    \centering
    \includegraphics[width=\columnwidth]{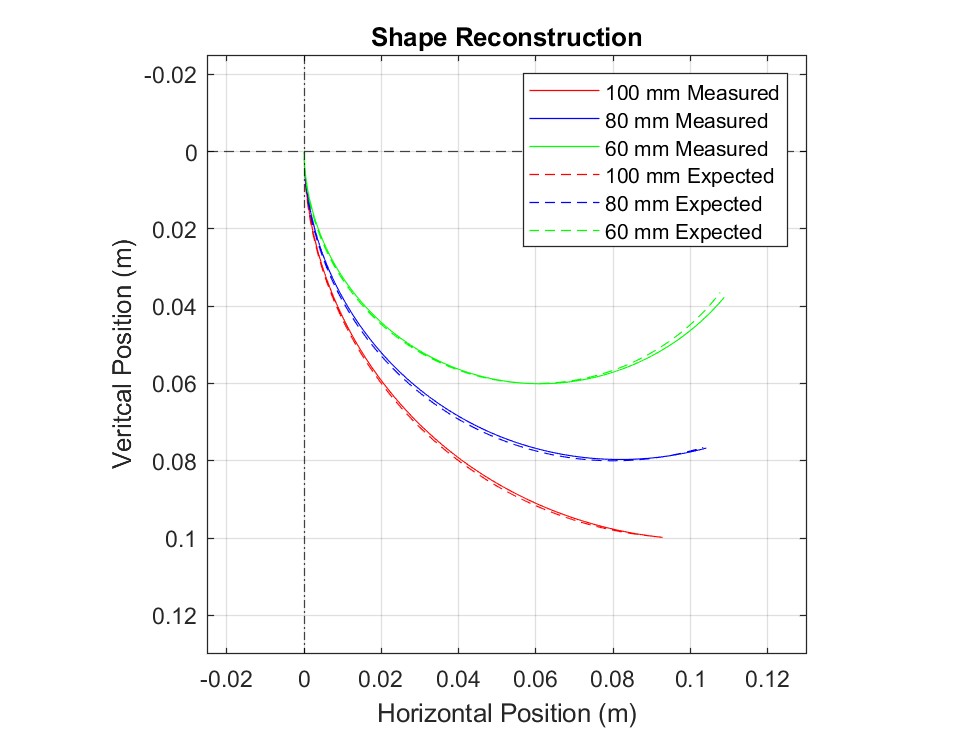}
    \caption{C-shape bending shape reconstruction. }
    \label{fig: C-Shape Shape Reconstruction}
\end{figure}

\begin{figure}[t]
    \centering
    \includegraphics[width=\columnwidth]{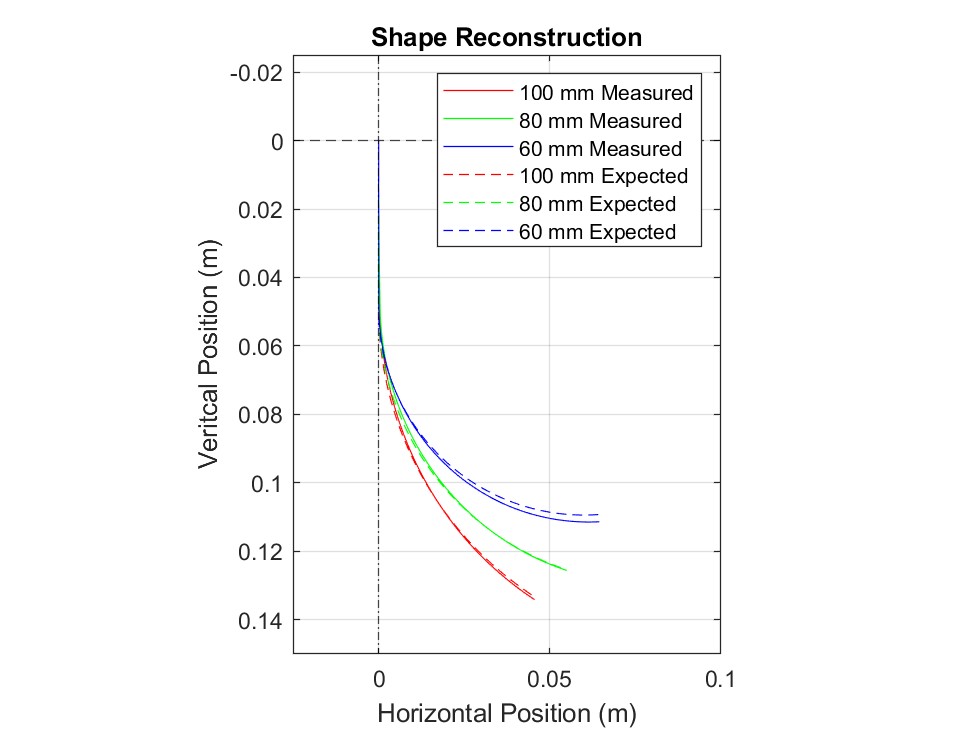}
    \caption{J-shape bending shape reconstruction. }
    \label{fig: J-Shape Shape Reconstruction}
\end{figure}

\subsection{C-Shape Bending Results }
\par The average strain and estimated radius of curvature of the performed SS in the C-shape jig  along the length of the fiber optic sensor  are displayed in Fig. \ref{fig: C-Shape Strain and Curvature Plots} for each trial. 
Also, all values for the C-shape bending experiment are shown in Table \ref{tab: C-Shape Data}. Both the average strain and radius were estimated along the entire length of the OFDR-SSA considered. The average radius of curvature per curve was used to solve for the estimated X-Y coordinates of the SSA and connect them to complete the shape reconstruction process shown in Fig. \ref{fig: C-Shape Shape Reconstruction}.

\begin{table*}[t]
\centering 
\caption{C-Shape Bending Results}
\begin{tabular}{lccccc}
\toprule
\textbf{Curvature} & \textbf{Average Strain (mS)}& \textbf{Average Radius (mm)}& \textbf{Tip Position Error (mm)}& \textbf{Shape Error (mm)}& \textbf{Average Area Error (mm\textsuperscript{2})}\\
\midrule
 C1 (100 mm)& 1440.649206& 100.871357& 2.097179& 1.166120& 1.476396\\
C2 (80 mm) & 1784.726425& 81.871782& 0.817171& 0.593618& 0.486622\\
C3 (60 mm) & 2397.258935& 61.460192& 1.638290& 0.489858& 1.155276\\
\bottomrule
\end{tabular}
\label{tab: C-Shape Data}
\end{table*}

\begin{table*}[t]
\centering
\caption{J-Shape Bending Results}
\begin{tabular}{lccccc}
\toprule
\textbf{Curvature} & \textbf{Average Strain (mS)}& \textbf{Average Radius (mm)}& \textbf{Tip Position Error (mm)}& \textbf{Shape Error (mm)}& \textbf{Average Area Error (mm\textsuperscript{2})}\\
\midrule
J1 (100 mm) & 935.08632 & 104.393193 & 1.677861 & 0.542728 & 1.176259 \\
J2 (80 mm) & 1176.915874 & 83.546379 & 1.041806 & 0.456734 & 0.686597 \\
J3 (60 mm) & 1511.251464 & 63.81685 & 3.36647 & 1.135024 & 2.31199 \\
\bottomrule
\end{tabular}
\label{tab:J-Shape Data}
\end{table*}

\subsection{J-Shape Bending Results}
The average strain and estimated radius of curvature of the performed SS in the J-shape jig  along the length of the fiber optic sensor  are displayed in Fig. \ref{fig: J-Shape Strain and Curvature Plots} for each trial. As shown in Fig. \ref{fig:C and J Shape Jigs}c, of the $150$ mm length of the CM, $50$ mm of the  length was within the first straight path of the J-curve. The remaining $100$ mm were placed within the curved part of the J-curve. This length was also used to calculate the average radius in Table\ref{tab:J-Shape Data}. Fig. \ref{fig: J-Shape Shape Reconstruction} also shows the reconstructed shape of SSA for the J-shape experiments.

\section{Discussion}

The reconstructed C-shapes shown in Fig. \ref{fig: C-Shape Shape Reconstruction}, clearly demonstrate the high accuracy of the proposed SSA and reconstruction process. Also, investigation of Table \ref{tab: C-Shape Data} shows that C2 has the smallest errors (i.e., 0.81 mm tip error, 0.59 mm shape error, and 0.48 $mm^2$ area error) compared to the other two curves. Among the C-shape experiments, C1 had the greatest tip and average area errors with approximately $2.1$ mm and $1.16$ $mm^2$, respectively, but C2 had the largest shape error of approximately $0.593618$ mm. It is worth emphasizing that considering the relatively long shape of CM (i.e., 170 mm), the obtained results are very acceptable compared with the literature. and,  particularly,  the study performed in \cite{monet2020high}. In that study, authors used a  CM with 35 mm length and compared performance of  FBG-based and OFDR-based SS. The obtained average (and max) SS error for such short length was reported as 0.5 mm (2.34 mm) and 0.28 mm (1.37 mm) mm shape deviation errors for FBG and OFDR SS.  It is worth noting that for this SS process, authors used 2 SSA with 3 FBG fibers on each SSA and two  OFDR sensors whereas here we are using a single fiber attached on a flat wire. 

 The reconstructed J-shapes shown in Fig. \ref{fig: C-Shape Shape Reconstruction}, clearly demonstrate the high accuracy of the proposed SSA and reconstruction process for the considered long CM. The values obtained in the J-shape experiment showed that the average radii for each curve were $104.393$ mm, $83.546$ mm, and $63.817$ mm when their respective expected results were $100$ mm, $80$ mm, and $ 60$ mm. Unlike  the C-shape bending experiments, the errors in this bending were consistent per curve. J2 had the smallest errors (i.e., 1.04 tip error, 0.45 mm shape error, and 0.68 $mm^2$ area error) while J3 had the greatest errors (i.e., 3.3 mm tip error, 1.13 mm shape error, and 2.32 $mm^2$ area error). Similar to the C-shape reconstructions, the obtained results are very acceptable considering the long length of the CM and using a SSA with single fiber. 
 
 Of note, the obtained phenomenal performance in reconstructing the C- and specifically J-shape trajectories  can be directly attributed to the distributed SS feature of OFDR compared with FBG fibers in which measurements are solely performed in limited amount of nodes. This great feature enables accurate shape reconstructions even when a sharp change in the measured strain is observed. Investigation of Fig.  \ref{fig: J-Shape Strain and Curvature Plots} clearly shows such a sudden change in the measured strain (where the straight segment meets the curved part of J-shape trajectory) that perfectly has been captured by the OFDR fiber. It is clear an FBG fiber will miss this drastic change in the strain if there are not enough number of nodes around the area in which this sharp change  happening.

\section{Conclusion}
In this paper, we featured a unique SSA utilizing a single OFDR fiber attached to a flat wire and demonstrated its phenomenal performance in SS of a long and soft CM with 170 mm length. Performance of this SSA was evaluated through 3 different C- and J-shape trajectories.  Results demonstrated phenomenal SS accuracy for both C-shape ($<$2 mm tip error, $<$1.2 mm shape error, and $<$1.5 $mm^2$ area error) and J-shape ($<$ 1.6 mm tip error, $<$1.2 mm shape error, and $<$2.3 $mm^2$ area error) experiments.  Overall, we demonstrated that the proposed SSA with single fiber, not only  facilitates the process of SSA fabrication, but it also results in a high SS accuracy using a simple reconstruction method.
In the future, we will evaluate performance of the proposed SSA in dynamic SS scenarios and inside various obstructed environments.


\bibliographystyle{unsrt}
\bibliography{main}
\end{document}